# LEVERAGING DEEP NEURAL NETWORK ACTIVATION ENTROPY TO COPE WITH UNSEEN DATA IN SPEECH RECOGNITION

*Vikramjit Mitra, Horacio Franco*

Speech Technology and Research Laboratory, SRI International, Menlo Park, CA, USA
{vikramjit.mitra, horacio.franco}@sri.com

## ABSTRACT

Unseen data conditions can inflict serious performance degradation on systems relying on supervised machine learning algorithms. Because data can often be unseen, and because traditional machine learning algorithms are trained in a supervised manner, unsupervised adaptation techniques must be used to adapt the model to the unseen data conditions. However, unsupervised adaptation is often challenging, as one must generate some hypothesis given a model and then use that hypothesis to bootstrap the model to the unseen data conditions. Unfortunately, reliability of such hypotheses is often poor, given the mismatch between the training and testing datasets. In such cases, a model hypothesis confidence measure enables performing data selection for the model adaptation. Underlying this approach is the fact that for unseen data conditions, data variability is introduced to the model, which the model propagates to its output decision, impacting decision reliability. In a fully connected network, this data variability is propagated as distortions from one layer to the next. This work aims to estimate the propagation of such distortion in the form of network activation entropy, which is measured over a short-time running window on the activation from each neuron of a given hidden layer, and these measurements are then used to compute summary entropy. This work demonstrates that such an entropy measure can help to select data for unsupervised model adaptation, resulting in performance gains in speech recognition tasks. Results from standard benchmark speech recognition tasks show that the proposed approach can alleviate the performance degradation experienced under unseen data conditions by iteratively adapting the model to the unseen data's acoustic condition.

*Index Terms*—automatic speech recognition, robust speech recognition, unsupervised adaptation, neural network activations, confidence measures.

## 1. INTRODUCTION

Deep learning technologies have become the preferred technique for building automatic speech recognition (ASR) systems [1, 2, 3], demonstrating impressive performance gains for almost all tried languages and acoustic conditions. Interestingly, deep neural network (DNN)-based systems are both data hungry and data sensitive [4], with model performance typically found to improve as more and more data from disparate conditions are used to train the model. Unfortunately, labeled data can be expensive. And although large volumes of data become available every day, not all of it is properly transcribed or reflective of the varying acoustic conditions that systems must tackle. With scarce data, DNN acoustic models are found to be quite sensitive to acoustic condition mismatches, where a subtle change in the background acoustic conditions due to noise, reverberation, and microphone conditions can significantly worsen model performance.

To combat such data-mismatch problems, multi-condition training accompanied by data augmentation is typically used to expose the DNN acoustic models to various possible background conditions. Multi-condition training was reported in [5], where a DNN acoustic model was found to benefit from training with thousands of hours of acoustic data collected from diverse sources. Data augmentation [6, 7] has been found to benefit ASR performance in reverberant conditions, where simulating different room impulse responses is relatively easy compared to simulating non-stationary background noise sources. Typically, data augmentation relies on artificially coloring the speech with additive noise or reverberation effects. In multi-condition training and data-augmentation approaches, prior knowledge is generally assumed about the kind of distortion the model will see, which often may not be true. Augmentation may expose the model (to some extent) to the anticipated acoustic variations; but in reality, acoustic variations are difficult to anticipate. Real-world ASR applications encounter diverse acoustic conditions, which are mostly unique and hence difficult to model. One such condition is reverberation and noise, which practically is an open-set problem. Systems that are trained with several thousands of hours of data collected from different realistic conditions typically are found to be quite robust to background conditions, as they are expected to contain a lot of variations; however, such data may not contain all the possible variations found in the world, and more specifically, such data may be available for some popular languages but not for others.

Recently, several open speech recognition evaluations (such as the MGB [8], CHiME [9], ASpIRE [10], and

REVERB [27] challenges) have shown how vulnerable DNN-hidden Markov model (HMM) acoustic models are to realistic, varying, and unseen acoustic conditions. One of the most celebrated and least resource-constrained approaches to coping with unseen data conditions is performing unsupervised adaptation, where the only necessity is having raw data. A more reliable adaptation technique is supervised adaptation, which assumes having some annotated target-domain data; however, annotated data is often unavailable in real-world scenarios. This constraint often makes unsupervised adaptation more practical.

Unsupervised speaker adaptation of DNNs has been explored with much success [11–13], with adaptation based on maximum likelihood linear regression (MLLR) transforms, i-vectors, etc. showing impressive performance gains over un-adapted models. In [4], stacked bottleneck (SBN) neural network architecture was proposed to cope with limited target-domain data, with the SBN net used as a feature extractor. The SBN system was used to handle unseen languages in [4] and, in [7], was extended to cope with unseen reverberation conditions. In [14], Kullback-Leibler divergence (KLD) regularization was proposed for DNN model parameter adaptation, which differs from the typically used L2 regularization [15] in the sense that it constrains the model parameters themselves rather than the output probabilities. Feature-space maximum likelihood linear regression (fMLLR) transformed feature was found in [13] to improve DNN acoustic model performance for mismatched cases. In [26], using confidence filtering was shown to improve acoustic performance.

In this work, we focus on understanding how acoustic condition mismatch between the training and the testing data impacts internal information flow within a fully connected neural network. Similar efforts have been pursued by researchers in [30, 32]. This paper investigates how data mismatch impacts the neural activations, and given that knowledge, this work investigates a way to use neural activations to predict when a DNN's decision may be less accurate. We use the neural activations to create a reliability measure for selecting untranscribed data for acoustic model adaptation. We have explored if such process can improve recognition performance in an unseen acoustic condition (reverberated speech) through iterative adaptation and if such process can generate performance as good as the seen acoustic conditions (i.e., when the model is trained with reverberated speech).

## 2. DATA

The acoustic models in this work were trained by using the multi-conditioned, noise- and channel-degraded training data from the 16 kHz Aurora-4 noisy *Wall Street Journal* (WSJ0) corpus. Aurora-4 contains a total of six additive noise types with channel-matched and mismatched conditions. It was created from the standard 5K WSJ0 database and contains 7180 training utterances of approximately 15-hours duration and 330 test utterances.

The Aurora-4 test data includes 14 test sets from two different channel conditions and six different added noises (car; babble; restaurant; street; airport; and train station) in addition to the clean condition. The signal-to-noise ratio (SNR) for the test sets varied between 0 and 15 dB. The evaluation set consists of 5K words under two different channel conditions. The original audio data for test conditions 1–7 was recorded with a Sennheiser microphone, while test conditions 8–14 were recorded using a second microphone randomly selected from a set of 18 different microphones [20]. Results from the evaluation set are presented as follows: Set A: clean, matched-channel (test set 1); Set B: noisy, matched-channel (test sets 2–7); Set C: clean, varying-channels (test set 8); and Set D: noisy, varying-channels (test sets 9–14).

In this work, we treated reverberation as the unseen data condition, training our models with the Aurora-4 corpus and assessing model performance on real-world reverberated data. For adaptation, optimization, and evaluation purposes, we used the training, development, and the evaluation sets distributed with the REVERB 2014 challenge, respectively. The REVERB 2014 speech dataset [27] contains single-speaker utterances, where only the single-microphone part of the dataset was used in the experiments reported in this paper. The REVERB 2014 training set consists of the clean WSJCAM0 [28] dataset, which was convolved with room impulse responses (with reverberation times from 0.1 sec to 0.8 sec) and then corrupted with background noise; hence, the training set consisted of artificially noise- and reverberation-corrupted data. Please note that as the REVERB 2104 training set was used as the unsupervised adaptation set, its transcriptions were not used in our experiments, except in the Oracle experiment, where the model was trained using both noisy and reverberated acoustic conditions. The evaluation and development data contain both real recordings (real data) and simulated data (sim data). The real data is borrowed from the MC-WSJ-AV corpus [29], which consists of utterances recorded in a noisy and reverberant room. The simulated evaluation set contained 1088 utterances in each of the far- and near-microphone conditions, each of which was split into three room conditions (1, 2, and 3). The real evaluation set contained 372 utterances split equally between far- and near-microphone conditions. Note that none of our experiments used speaker-level information.

## 3. ACOUSTIC FEATURES

We used gammatone filterbank energies (GFBs) as the acoustic features for our experiments. In the GFB processing, the input speech signal was analyzed by using a bank of 40 gammatone filters equally spaced on the equivalent rectangular bandwidth (ERB) scale. Within an analysis window of approximately 26 ms, the power of the bandlimited time signals was computed at a frame rate of 10 ms. The subband powers were root-compressed by using the

15th root, and the resulting 40-dimensional feature vector was used as the GFBs.

## 4. THE ENTROPY MEASURE

The drawback of existing supervised learning approaches is that the resulting models only learn the information that is present in the training set. When faced with unknown variations, such models fail to generalize well and consequently propagate any distortion in the input features, resulting in distorted outputs that do not represent relevant aspects of the input [30, 32]. In a fully connected DNN, such anomalies propagate from one layer to the next and, like a ripple effect, spread localized distortions across all dimensions of the neural network's hidden layers. Techniques such as drop-out training usually help in such cases, as they reduce the reliability of one neuron over the other and help the model improve its generalization capability.

In grossly mismatched situations, detecting the test cases that cause the system to completely fail, versus those that generate a reasonable output can be quite useful. One way to generate such detection is through a confidence measure, which is generally indicative of how trustworthy the ASR hypothesis is for each of the test files.

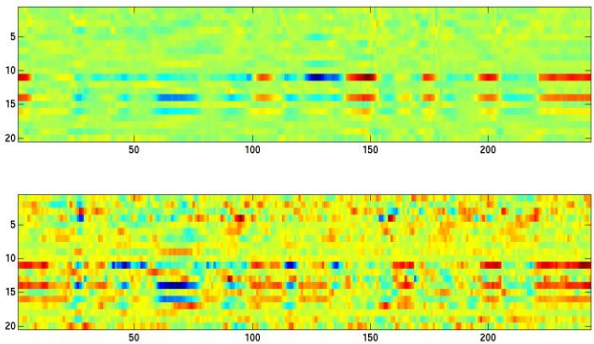

Figure 1. Image of DNN activations from a hidden layer (activations from only 20 neurons are shown here) for seen [top] and unseen [bottom] acoustic data. The abscissa gives the time steps in units, where each unit is 10 ms apart.

A fully connected network can be interpreted as a cascade of several feature-transformation steps, where the goal is making each target class as discriminative as possible with respect to each other. Hence, for cases where the model fails to generate reasonable performance, such transformations fail to generate reliable features, and therefore the model decision is impacted. The veracity of the above statement is observed in the Figure 1, where we show the neural activations generated from seen versus unseen data, both corrupted with noise.

If we consider a hidden layer $N$ having $n$ neurons each, generating activations $x_{t,i}$ at a given instant of time $t$, for $i^{th}$ neuron, then we can estimate the entropy of the activation (after soft-max) of the $i^{th}$ neuron over a time window of $m$ centered around $t$. Let us assume $X_{t,i}$ as a random vector representing the activations of neuron $i$ at hidden layer $N$, over a time window $m$ centered around $t$.

$$X_{t,i} = \left[x_{t-\frac{m}{2}}, x_{t-\frac{m}{2}+1}, \ldots x_t, \ldots x_{t+\frac{m}{2}}\right]$$

then the entropy of $X_{t,i}$ is defined as

$$H_{X_{t,i}} = -\sum p(X_{t,i}) \log [p(X_{t,i})] \quad (1)$$

where $p(X_{t,i})$ is the probability density function of $X_{t,i}$. Note that according to (1), the entropy is obtained for each activation $i$ over a running window of $m$. In our experiments, we used an $m$ of 91 samples (~1 sec) centered at time instant $t$. We used a frame hopping of 20 frames (~50 ms) to estimate the entropy (i.e., the analysis point $t$ was moved at a rate of 20 frames from each other). Note that the selection of the value of $m$ was done by maximizing the correlation of the run-time estimated entropy measure with the observed word error rate (WER) from the Aurora-4 test set.

Finally, once the run-time entropy was obtained from each of the $n$ neurons in the $N^{th}$ hidden layer, a summary measure was obtained, by estimating the mean entropy for each neuron activation, resulting in a vector of dimension equal to the number of neurons in that layer. This vector was then sorted, the top 70th percentile entropy measures across the activations were selected, and their mean value was computed to generate the final normalized and ranked summary entropy measure (NRSE). Note that the NRSE measure is a single real number computed for each audio file present in the adaptation dataset. It was observed that the estimated entropy correlated with ASR WER with a correlation coefficient of approximately 0.4 for an unseen noisy speech dataset. Given this, the natural question that follows is from which layer should the activation be used to estimate NRSE (i.e., "what should be the value of $N$?"), a question which we explored in our speech recognition studies presented in this paper.

## 5. ACOUSTIC MODEL

In earlier work [25], we showed that CNN models perform much better than DNN acoustic models for the Aurora-4 speech recognition task. Typically, CNNs give lower WERs compared to DNNs when using filterbank features for ASR tasks, and GFBs perform better or as well as the mel-filterbank energies (MFBs). To generate the alignments necessary for training the CNN system, a GMM-HMM model was used to produce senone labels. Altogether, the GMM-HMM system produced 3125 context-dependent (CD) states for the Aurora-4 training data. The input features to the acoustic models were formed by using a context window of 15 frames (7 frames on either side of the current frame).

The acoustic models were trained by using cross-entropy on the alignments from the GMM-HMM system. A five hidden layered DNN with 2048 neurons in each layer was trained by using the alignments from the GMM-HMM system, which in turn was used to generate alignments for training the subsequent DNN/CNN acoustic models trained in this paper.

The CNN acoustic models consisted of 200 filters of size 8, and the resulting feature maps were sub-sampled using max-pooling over three samples without overlap. The subsequent fully connected network had four hidden layers, with 2048 neurons per hidden layer, and the output layer included as many nodes as the number of CD states for the given dataset. The networks were trained by using an initial four iterations with a constant learning rate of 0.008, followed by learning rate halving based on cross-validation error decrease. Training stopped when no further significant reduction in cross-validation error was noted or when cross-validation error started to increase. Backpropagation was performed by using stochastic gradient descent with a mini-batch of 256 training examples. We observed that time-frequency convolution (using TFCNN architecture) performed better than the one-dimensional frequency convolution typically done in CNN acoustic models [25], and hence in almost all of our experiments, we used the TFCNN acoustic model to report our findings.

The TFCNN architecture is similar to [25], where two parallel convolutional layers are used at the input, one performing convolution across time, and the other across the frequency axis of the input filterbank features. For the TFCNN acoustic models, the input acoustic features were formed by using a context window of 17 frames (8 frames on either side of the current frame). The TFCNNs had 75 filters to perform time convolution and 200 filters to perform frequency convolution. For time and frequency convolution, eight bands were used, followed by a max-pooling over three samples after frequency convolution, and a max-pooling over five samples for time convolution. The feature maps after both the convolution operations were concatenated and then fed to a fully connected neural net, which had 2048 nodes and four hidden layers.

## 6. RESULTS

The baseline acoustic model was trained with Aurora-4 multi-condition training dataset, where a held-out cross-validation set was used to train the neural net acoustic models, similar to [25]. The reverberated acoustic condition is treated as the unseen data condition in our experiments, where the experimental analysis was performed by using the development and test data from the REVERB 2014 challenge dataset. As an Oracle experiment we trained a $CNN_{ORACLE}$ and a $TFCNN_{ORACLE}$ system, which were trained jointly with Aurora-4 and REVERB 2014 training data. The WER results from the baseline CNN and TFCNN systems (trained only with Aurora-4) and the $TFCNN_{ORACLE}$ system are shown in Tables 1, 2, and 3 for the Aurora-4 test set, and the REVERB 2014 dev and test sets, respectively.

Table 1 shows that for all the conditions, the TFCNN acoustic model performed better than its CNN counterpart, and interestingly that augmenting the training data with additional reverberated data improved the performance on the Aurora-4 test set, even if that test set did not have any reverberation in them, indicating the benefit of data augmentation, which has been reported several times by other studies [5, 6, 7].

Table 1. WERs from different acoustic models when evaluated on the Aurora-4 test set.

| System | Aurora-4 | | | | |
|---|---|---|---|---|---|
| | A | B | C | D | Avg. |
| CNN | 2.8 | 6.0 | 5.7 | 14.7 | 9.5 |
| TFCNN | 2.9 | 5.7 | 5.6 | 14.3 | 9.2 |
| $CNN_{ORACLE}$ | 2.9 | 5.6 | 5.5 | 14.0 | 9.0 |
| $TFCNN_{ORACLE}$ | 2.6 | 5.5 | 5.4 | 13.8 | 8.8 |

Table 2. WERs from different acoustic models when evaluated on the REVERB 2014 dev set.

| System | REVERB 2014 dev | |
|---|---|---|
| | Avg. Sim | Avg. Real |
| CNN | 41.3 | 43.9 |
| TFCNN | 39.3 | 42.4 |
| $CNN_{ORACLE}$ | 10.9 | 21.7 |
| $TFCNN_{ORACLE}$ | 10.3 | 21.3 |

Table 3. WERs from different acoustic models when evaluated on the REVERB 2014 test set.

| System | REVERB 2014 test | |
|---|---|---|
| | Avg. Sim | Avg. Real |
| CNN | 38.1 | 47.8 |
| TFCNN | 37.8 | 46.9 |
| $CNN_{ORACLE}$ | 10.5 | 22.6 |
| $TFCNN_{ORACLE}$ | 10.0 | 21.2 |

Tables 2 and 3 reflect the impact of unseen data conditions, where the performance of the acoustic model was found to degrade by more than 50% relative for unseen reverberated condition as opposed to the seen reverberated condition, as reflected by the Oracle experiments. In all cases, the TFCNNs were found to perform slightly better than their CNN counterparts.

Based on the observations shown in Tables 1 to 3, we focused on the TFCNN acoustic models for our subsequent experimental evaluations. To evaluate the relevance of the NRSE measure and which layer of the TFCNN acoustic model they should be derived from, we treated the REVERB ,2014 training set as the unsupervised adaptation dataset and performed the following experiments: First, we evaluated the case where the entire REVERB 2014 training set decoded hypothesis from the TFCNN acoustic model was used to generate alignments for adapting the TFCNN acoustic model trained with Aurora-4 data. The resulting adapted TFCNN acoustic model was treated as the baseline system, and we named it as $TFCNN_{ALL\_P0}$, where ALL reflects that the entire REVERB 2014 hypothesis was used to adapt the model, and P0 reflects that this was the first pass on unsupervised adaptation. Next, we extracted the activations from hidden layers 2–5 of the TFCNN acoustic model, from which we estimated the NRSE measures and used that to select the top 4K segments in each case, which were found to have the lower entropy. Tables 4, 5, and 6 reflect the results from the TFCNN acoustic model after adaptation on the Aurora-4 test set, and the REVERB 2014 dev and test sets, respectively. Note that during adaptation, the unsupervised adaptation dataset was used in addition to the original Aurora-4 training dataset to update the acoustic model parameters. During adaptation, all model parameters were updated with an $L_2$ norm of 0.001 and an initial learning rate of 0.004, with the learning rate halved at every step. Early stopping was performed based on the cross-validation error where the Aurora-4 cross validation set was used. Note that we did not use any reverberated data for cross-validation purposes.

Table 4. WERs from adapted acoustic models when evaluated on the Aurora-4 test set.

| System | Aurora-4 | | | | |
|---|---|---|---|---|---|
| | A | B | C | D | Avg. |
| $TFCNN_{ALL\_P0}$ | 3.2 | 5.9 | 6.2 | 14.7 | 9.5 |
| $TFCNN_{L5\_P0}$ | 3.3 | 5.8 | 6.1 | 14.6 | 9.4 |
| $TFCNN_{L4\_P0}$ | 3.2 | 5.8 | 5.8 | 14.5 | **9.3** |
| $TFCNN_{L3\_P0}$ | 3.3 | 6.0 | 6.0 | 14.7 | 9.5 |
| $TFCNN_{L2\_P0}$ | 3.1 | 5.9 | 6.2 | 14.7 | 9.5 |

Table 5. WERs from different acoustic models when evaluated on the REVERB 2014 dev set.

| System | REVERB 2014 dev | |
|---|---|---|
| | Avg. Sim | Avg. Real |
| $TFCNN_{ALL\_P0}$ | 24.4 | 33.7 |
| $TFCNN_{L5\_P0}$ | 24.0 | 33.3 |
| $TFCNN_{L4\_P0}$ | 23.3 | 32.8 |
| $TFCNN_{L3\_P0}$ | **23.0** | **32.2** |
| $TFCNN_{L2\_P0}$ | 23.1 | 32.8 |

Table 6. WERs from different acoustic models when evaluated on the REVERB 2014 test set.

| System | REVERB 2014 test | |
|---|---|---|
| | Avg. Sim | Avg. Real |
| $TFCNN_{ALL\_P0}$ | 22.7 | 37.4 |
| $TFCNN_{L5\_P0}$ | 22.3 | 36.4 |
| $TFCNN_{L4\_P0}$ | 21.7 | 36.1 |
| $TFCNN_{L3\_P0}$ | 21.5 | 36.4 |
| $TFCNN_{L2\_P0}$ | 21.5 | 36.6 |

Tables 4, 5, and 6 show that data selection followed by model adaptation obtained better performance than using the entire adaptation data. This indicates that the data-selection process helps prune some bad hypothesis from the parent recognition system. Comparing Tables 1 and 4, we can see that including the Aurora-4 training data in the adaptation set helped the model to retain its performance on noisy conditions, while improving its performance significantly on the reverberated speech conditions. From Tables 5 and 6, we can state that NRSE measures from layer 3 (i.e., one of the intermediate layers in the network) were a better choice for data selection. This observation could be related to the role of the intermediate layers, which are known to perform different feature transformations, resulting in more discriminative features to aid the decision-making task of the final layers. Seemingly, the entropies of these intermediate layers are likely correlated with the word error rates, as higher distortions in the feature space may be contributing to more confusion in the decision-making task of the final layers.

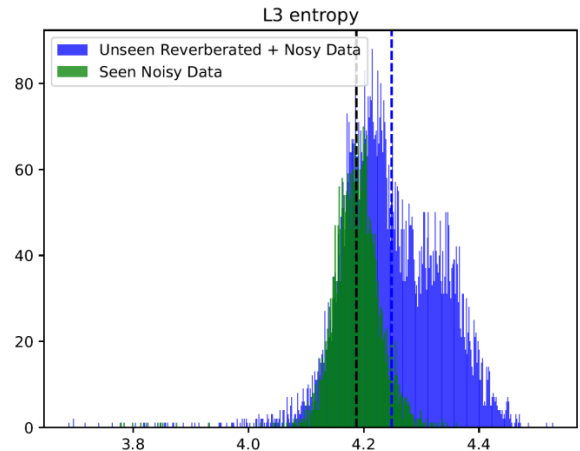

Figure 2. Distribution of NRSE from activations obtained from the third hidden layer of the TFCNN acoustic model trained with Aurora-4 multi-conditioned data. **Green**: NRSE from the Aurora-4 training data. **Blue**: NRSE from the unseen reverberated + noisy data (in this case, the REVER 2014 training data). The vertical dotted lines indicate their respective means.

Comparing Tables 2 and 3 with Tables 5 and 6 shows that the unsupervised adaptation using data selection significantly reduced the WER. For the simulated reverberation condition, the relative WER improvement was quite significant (i.e., greater than 40%); whereas for the real reverberation conditions, the improvements were greater than 20% for both the dev and test sets, respectively. The significant improvement on the simulated reverberation condition is to some extent expected, as the adaptation set used in this case was the REVERB 2014 training set, which consists of simulated reverberation only; hence, it helped the model to learn that condition more than the real reverberation condition.

To analyze if recognition performance can be improved further through subsequent adaptation of the acoustic model, we performed multiple passes of adaptation, where the hypothesis (of the adaptation set) after each adaptation step was used to adapt the acoustic model, which we continued for two more steps. The goal was investigating if, through multiple iteration of adaptation, we can further improve the performance of the model.

Table 7 shows that through iterative adaptation (where the number of files during data selection was increased by 1000, starting from the 4000 files selected in the first step), the model improved its performance on the reverberated conditions, with 16% and 14% relative reduction in WER was obtained from the third-pass adapted model (TFCNN$_{L3\_P2}$) compared to the first-pass adapted model (TFCNN$_{L3\_P0}$). Interestingly, the fourth-pass adapted model (TFCNN$_{L3\_P3}$) did not show a significant gain over the third-pass model (TFCNN$_{L3\_P2}$).

Table 7. WERs from the mismatched model, Oracle model, and adapted model (after multiple passes (P) from 0 to 3) when evaluated on the REVERB 2014 test set.

| System | REVERB 2014 test | |
|---|---|---|
| | Avg. Sim | Avg. Real |
| TFCNN | 37.8 | 46.9 |
| TFCNN$_{ORACLE}$ | 10.0 | 21.2 |
| TFCNN$_{L3\_P0}$ | 21.5 | 36.4 |
| TFCNN$_{L3\_P1}$ | 19.2 | 34.0 |
| TFCNN$_{L3\_P2}$ | 18.1 | 31.3 |
| TFCNN$_{L3\_P3}$ | 17.7 | 31.0 |

## 7. CONCLUSION

In this work, we investigated whether activations from neural net hidden layers can be used to predict the reliability of the neural net's decision, and hence use that prediction to perform data selection for doing unsupervised model adaptation. In a fully connected network, information flows from left to right, and if unseen distortions are introduced through the input, they will propagate through the hidden layers to the output nodes. When such distortions are propagated, they should be detectable at the hidden layers. And if they are detected, they can inform whether the network is likely to generate a reliable decision versus a less reliable one. Based on such assumptions, we proposed a new measure: the normalized and ranked summary entropy (NRSE) measure, which estimates the overall percentile entropy of the neural net's activation for a given input segment. We observed that a high NRSE indicates that the network is more likely to generate an erroneous hypothesis, compared to a lower NRSE, which indicates that the hypothesis is more likely to be a reliable one. Based on this, we explored data selection for unsupervised model adaptation and demonstrated that the data-selection process was helpful for unsupervised model adaptation, reducing ASR error rates for unseen data conditions. In addition, we found that performing multiple passes of unsupervised adaptation resulted in further improvement in recognition performance, reducing the performance gap between acoustic models trained with seen and unseen acoustic conditions.

In this work, we performed data selection through a rank-sorting of NRSE measures estimated from each audio file present in the unsupervised dataset. A more pragmatic approach would be to perform data selection through thresholding of the NRSE measures, a direction we plan to explore in the future. In addition, future studies should also explore applying the proposed approach to other datasets and investigate ways to perform continuous adaptation.

## 10. ACKNOWLEDGEMENTS

This material is based upon work partly supported by the Defense Advanced Research Projects Agency (DARPA) under Contract No. HR0011-15-C-0037. The views, opinions, and/or findings contained in this article are those of the authors and should not be interpreted as representing the official views or policies of the Department of Defense or the U.S. Government.

## 11. REFERENCES

[1] A. Mohamed, G.E. Dahl, and G. Hinton, "Acoustic modeling using deep belief networks," *IEEE Trans. on ASLP*, vol. 20, no. 1, pp. 14–22, 2012.

[2] F. Seide, G. Li, and D. Yu, "Conversational speech transcription using context-dependent deep neural networks," *Proc. of Interspeech*, 2011.

[3] G. Hinton, L. Deng, D. Yu, G. Dahl, A.-r. Mohamed, N. Jaitly, A. Senior, V. Vanhoucke, P. Nguyen, T. Sainath, and B. Kinsgbury, "Deep neural networks for acoustic modeling in speech recognition," *IEEE Signal Process. Mag.*, vol. 29, no. 6, pp. 82–97, 2012.

[4] F. Grézl, E. Egorova, and M. Karafiát, "Further investigation into multilingual training and adaptation of stacked bottle-neck neural network structure," *Proc. of SLT*, pp. 48–53, 2014.

[5] T. Sainath, R.J. Weiss, K. Wilson, A.W. Senior, and O. Vinyals, "Learning the speech front-end with raw waveform CLDNNs," *Proc. of Interspeech*, 2015.

[6] V. Peddinti, G. Chen, V. Manohar, T. Ko, D. Povey, and S. Khudanpur, "JHU ASpIRE system: Robust LVCSR with TDNNs, i-vector adaptation and RNN-LMS," *Proc. of ASRU*, 2015.

[7] M. Karafiát, F. Grézl, L. Burget, I. Szöke, and J. Cernocký "Three ways to adapt a CTS recognizer to unseen reverberated speech in BUT system for the ASpIRE challenge," *Proc. of Interspeech*, pp. 2454–2458, 2015.

[8] P. Bell, M.J.F. Gales, T. Hain, J. Kilgour, P. Lanchantin, X. Liu, A. McParland, S. Renals, O. Saz, M. Wester, and P.C. Woodland, "The MGB challenge: Evaluating multi-genre broadcast media recognition," *Proc. of ASRU*, 2015.

[9] J. Barker, R. Marxer, E. Vincent, and S. Watanabe, "The third 'CHiME' speech separation and recognition challenge: Dataset, task and baselines," *Proc. of ASRU*, 2015.

[10] M. Harper, "The automatic speech recognition in reverberant environments (ASpIRE) challenge," *Proc. of ASRU*, 2015.

[11] T. Yoshioka, A. Ragni, and M.J.F. Gales, "Investigation of unsupervised adaptation of DNN acoustic

models with filterbank input," *Proc. of ICASSP*, pp. 6344–6348, 2014.

[12] G. Saon, H. Soltau, D. Nahamoo, and M. Picheny, "Speaker adaptation of neural network acoustic models using i-vectors," *Proc. of ASRU*, pp. 55–59, 2013.

[13] S.H.K. Parthasarathi, B. Hoffmeister, S. Matsoukas, A. Mandal, N. Strom, and S. Garimella, "fMLLR based feature-space speaker adaptation of DNN acoustic models," *Proc. of Interspeech*, 2015.

[14] D. Yu, K. Yao, H. Su, G. Li, and F. Seide, "KL-divergence regularized deep neural network adaptation for improved large vocabulary speech recognition," *Proc. of ICASSP*, 2013.

[15] X. Li and J. Bilmes, "Regularized adaptation of discriminative classifiers," *Proc. ICASSP'06*, 2006.

[16] K. Walker and S. Strassel, "The RATS radio traffic collection system," *Proc. of Odyssey 2012-The Speaker and Language Recognition Workshop*, 2012.

[17] A. Stolcke, "SRILM—An extensible language modeling toolkit," *Proc. of ICSLP*, pp. 901–904, 2002.

[18] A. Venkataraman and W. Wang, "Techniques for effective vocabulary selection," *Proc. Eighth European Conference on Speech Communication and Technology*, pp. 245–248, 2003.

[19] A. Mandal, J. van Hout, Y-C. Tam, V. Mitra, Y. Lei, J. Zheng, D. Vergyri, L. Ferrer, M. Graciarena, A. Kathol, and H. Franco, "Strategies for high accuracy keyword detection in noisy channels," *Proc. of Interspeech*, pp. 15–19, 2013.

[20] V. Mitra, H. Franco, M. Graciarena, and A. Mandal, "Normalized amplitude modulation features for large vocabulary noise-robust speech recognition," *Proc. of ICASSP*, pp. 4117–4120, 2012.

[21] V. Mitra, H. Franco, and M. Graciarena, "Damped oscillator cepstral coefficients for robust speech recognition," *Proc. of Interspeech*, pp. 886–890, 2013.

[22] T. Ng, R. Hsiao, L. Zhang, D. Karakos, S.H. Mallidi, M. Karafiat, K. Vesely, I. Szoke, B. Zhang, L. Nguyen, and R. Schwartz, "Progress in the BBN keyword search system for the DARPA RATS program," *Proc. of Interspeech*, pp. 959–963, 2014.

[23] V. Mitra, W. Wang, H. Franco, Y. Lei, C. Bartels, and M. Graciarena, "Evaluating robust features on deep neural networks for speech recognition in noisy and channel mismatched conditions," *Proc. of Interspeech*, pp. 895–899, Singapore, 2014.

[24] J. Gehring, Y. Miao, F. Metze, and A. Waibel, "Extracting deep bottleneck features using stacked auto-encoders," *Proc. of ICASSP*, 2013.

[25] V. Mitra, and H. Franco, "Time-frequency convolution networks for robust speech recognition," *Proc. of ASRU*, 2015.

[26] Thomas Drugman, Janne Pylkkönen, and Reinhard Kneser, "Active and semi-supervised learning in ASR: Benefits on the acoustic and language models," *Proc. of Interspeech*, 2016.

[27] K. Kinoshita, M. Delcroix, T. Yoshioka, T. Nakatani, E. Habets, R. Haeb-Umbach, V. Leutnant, A. Sehr, W. Kellermann, R. Maas, S. Gannot, and B. Raj, "The REVERB Challenge: A common evaluation framework for dereverberation and recognition of reverberant speech," *Proc. of IEEE Workshop on Applications of Signal Processing to Audio and Acoustics (WASPAA)*, 2013.

[28] T. Robinson, J. Fransen, D. Pye, J. Foote, and S. Renals, "WSJCAM0: A British English speech corpus for large vocabulary continuous speech recognition," *Proc. ICASSP*, pp. 81–84, 1995.

[29] M. Lincoln, I. McCowan, J. Vepa, and H.K. Maganti, "The multi-channel *Wall Street Journal* audio visual corpus (MC-WSJ-AV): Specification and initial experiments," *Proc. of IEEE Workshop on Automatic Speech Recognition and Understanding*, 2005.

[30] H. Hermansky, L. Burget, J. Cohen, E. Dupoux, N. Feldman, J. Godfrey, S. Khudanpur, M. Maciejewski, S.H. Mallidi, A. Menon, T. Ogawa, V. Peddinti, R. Rose, R. Stern, M. Wiesner, and K. Veselý, "Towards machines that know when they do not know: Summary of work done at 2014 Frederick Jelinek Memorial Workshop," *Proc. of ICASSP*, pp. 5009–5013, 2015.

[31] V. Mitra, J. van Hout, W. Wang, C. Bartels, H. Franco, D. Vergyri, "Fusion strategies for robust speech recognition and keyword spotting for channel- and noise-degraded speech," in *Proc. of Interspeech*, 2016.

[32] H. Zaragoza, and d'A-B. Florence, "Confidence measures for neural network classifiers," *Proc. of the 7$^{th}$ Int. Conf. Information Processing and Management of Uncertainty in Knowledge Based Systems*, 1998.